\def\year{2019}\relax
\documentclass[letterpaper]{article} 
\usepackage{aaai19}  
\usepackage{times}  
\usepackage{helvet}  
\usepackage{courier}  
\usepackage{url}  
\usepackage{graphicx}  
\usepackage{bm} 
\usepackage{amsmath,amssymb,amsthm} 
\usepackage[linesnumbered,ruled,vlined]{algorithm2e}
\begin{document}
%
\title{Deep Bayesian Optimization on Attributed Graphs}
\author{Jiaxu~Cui,\textsuperscript{$1$$,$$2$}
Bo~Yang,\textsuperscript{$1$$,$$2$}\thanks{Corresponding author. ybo@jlu.edu.cn}
Xia~Hu\textsuperscript{$3$}\\
\textsuperscript{$1$}College of Computer Science and Technology, Jilin University, Changchun, China\\
\textsuperscript{$2$}Key Laboratory of Symbolic Computation and Knowledge Engineering of Ministry of Education, China\\
\textsuperscript{$3$}Department of Computer Science and Engineering, Texas A\&M University, College Station, United States}

\maketitle
\begin{abstract}
Attributed graphs, which contain rich contextual features beyond just network structure, are ubiquitous and have been observed to benefit
various network analytics applications. Graph structure optimization, aiming to find the optimal graphs in terms of some specific measures,
has become an effective computational tool in complex network analysis. However,
traditional model-free methods suffer from the expensive computational cost of evaluating graphs;
existing vectorial Bayesian optimization methods
cannot be directly applied to attributed graphs and have the scalability issue due to the use of Gaussian processes (GPs).
To bridge the gap, in this paper, we propose a novel scalable
Deep Graph Bayesian Optimization (DGBO) method on attributed graphs. The proposed DGBO
prevents the cubical complexity of the GPs by adopting a
deep graph neural network to surrogate black-box functions,
and can scale linearly with the number of observations. Intensive experiments are conducted on both artificial and real-world problems, including
molecular discovery and urban road network design, and demonstrate the effectiveness of the DGBO compared with the state-of-the-art.
\end{abstract}

\section{1 ~ Introduction}
\begin{figure}[!htb]
\centering
\includegraphics[scale=0.54]{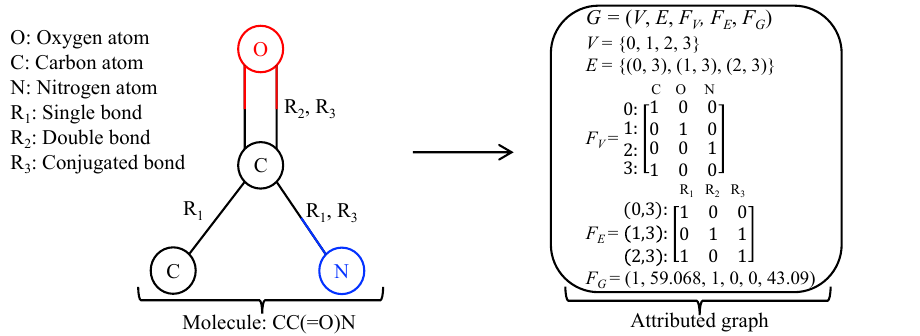}
\caption{An illustration from a molecule sampling in Delaney data set \cite{Delaney2004ESOL} to an attributed graph (defined in Section 2). Node feature is atomic type, edge feature is chemical bond type, and global attributes contain minimum degree, molecular weight, \# h-bond donors, \# rings, \# rotatable bonds, and polar surface area.}
\label{fig:mol2graph}
\end{figure}

Graphs have been intensively used to model network data generated in important application domains such as chemistry, transportation, social networks, and knowledge graphs. These real-world networks are often associated with a rich set of available
attributes with respect to nodes, edges, and global structures, which are known as attributed graphs. Fig.\ref{fig:mol2graph} provides one example, in which atomic type, chemical bond type, molecular weight,
polar surface area, and other attributes are observed on each molecule.
It has been studied that the attributes on graphs are highly correlated
to topological structures \cite{Zhang2013Citation} and
can benefit various network analysis tasks such as trust prediction
\cite{Tang2013Exploiting} and network embedding \cite{huang2017accelerated}.
Motivated by these observations, in this work we propose
to study whether the attributes on graphs can benefit the task of graph structure optimization and how to comprehensively explore available attributes to address this task more efficiently and effectively.

Optimizing the graph structure is a fundamental task in network analysis, aiming to find the optimal graphs with respect to some specific measures. Examples include discovering molecular structures with desired properties and designing road networks with better traffic conditions.
In the literature, existing graph structure optimization methods generally fall into two categories, i.e., model-free methods and model-based methods.

The model-free methods based on evolutionary strategies or annealing strategies have been widely applied to road network design \cite{Xiong1992Transportation,Miandoabchi2011Optimizing,Farahani2013A} and molecular discovery \cite{Supady2015First,Rupakheti2015A}. However, model-free methods will be less effective if the computational cost of evaluating graphs is expensive, because such methods usually require a large number of evaluations to maintain population diversity in finding an optimal solution. This is not acceptable particularly for the tasks with large-scale search spaces. For instance, to evaluate the effectiveness of a candidate molecular structure, one has to do lots of computer-aided simulations involving massive computing resources or do actual chemical experiments many times with high costs and potential risks.

Bayesian optimization (BO) \cite{DBLP:journals/pieee/ShahriariSWAF16}, a model-based global optimization framework, has shown its effectiveness in addressing the above-mentioned challenges. BO is particularly proposed to optimize black-box functions that are derivative-free, noisy, and expensive to evaluate with respect to different kinds of resources such as time and energy. Note that, for many graph structure optimization tasks, the objective functions are also black-box. In other words, we do not exactly know the mapping mechanism from structural space to measure space, which determines how the structure of a network will affect its functions or dynamics.

Unfortunately, existing BO algorithms focus on optimizing the objectives with vectorial inputs, such as hyper-parameter optimization \cite{DBLP:conf/nips/SnoekLA12} and robot control \cite{Cully:2015aa}. These methods cannot be directly applied to graphs or particularly attributed graphs mainly because 1) graph search space is non-Euclidean, discrete, and usually huge \cite{Polishchuk2013Estimation}, and 2) it is difficult for existing BO methods to properly and automatically extract the task-specific features from attributed graphs.

Some efforts have been devoted to employing BO, implicitly or explicitly, in optimizing graph structures. For examples, some work \cite{Dalibard2017BOAT,gardner17a} have been proposed to handle simple structures, not arbitrary, of vectorial components, which describe the predefined restrictions among them. Some others \cite{Kandasamy2018,Ramachandram2018Bayesian,Jin2018} were proposed to search optimal neural network architectures, by defining novel graph kernels to measure the similarities among neural networks. Note that these algorithms are exclusively designed for the task of neural architecture search and cannot be easily extended and applied to other domains.

Very recently, a BO framework was proposed explicitly for graph structure optimization \cite{Cui2018}, referred to as GBO (graph BO). In GBO, both structure and global attributes of the graph are considered to promote optimization performance by subtly combining deep graph kernels with vectorial kernels. According to their reports, GBO outperforms model-free methods and the BO methods with only graph kernels. Note GBO cannot be naturally extended to attributed graphs, where, except global attributes, the attributes of nodes and edges should also be considered.

More importantly, the aforementioned BO methods are mainly based on GPs (Gaussian processes). Although GPs is a very flexible non-parametric model for surrogating unknown functions and can effectively model uncertainty as well, the problem is the time cost of GPs inference grows cubically with the number of observations, as it necessitates the inversion of a dense covariance matrix. When search space becomes huge, a larger number of evaluations are needed to find optimal solutions, and thus GPs-based BO will be infeasible due to its cubic scaling.

To address the above challenges, in this paper, we study the problem of efficiently finding optimal attributed graphs. Specifically, we aim at answering two key questions: 1) How to properly represent attributed graphs and make full use of all available features to assist optimization process? and 2) How to make the optimization process more efficient and scalable? By investigating these questions, we propose a novel global Deep Graph Bayesian Optimization (DGBO) framework on attributed graphs. The key idea is to use a novel deep graph neural network to surrogate black-box functions instead of GPs.
The main contributions of this work are summarized as follows.
\begin{itemize}
\item Our proposed DGBO can make full use of available features to benefit graph structure optimization, and scales linearly with the number of observations.
\item The efficacy of the DGBO has been strictly validated on both artificial and real-world problems, which shows it effectively and efficiently handles large-scale problems.
\end{itemize}

\section{2 ~  Problem Statement}
\label{Problem_Statement}
The graph structure optimization we studied in this paper is described as: given a graph search space $\mathcal{G}$ and a task-specific expensive-to-evaluate black-box function $f:\mathcal{G} \to \mathbb{R}$, we aim at finding the optimal graph $G^* \in \mathcal{G}$ with the maximum value of $f$ at as low cost as possible. Mathematically, this problem is defined as:
\begin{equation}
	G^*=\mathop{\arg\max}_{G\in{\mathcal{G}}}[f( G )+\epsilon],
	\label{tab:graph_structure_optimization_problems}
\end{equation}
where  $\epsilon$ is the noise of evaluations, $G$ denotes an attributed graph defined as follows (see Fig.\ref{fig:mol2graph} for an example).

\noindent \textbf{Definition 1.}
\emph{(Attributed graph). $G=(V,E,F_V,F_E,F_G)$ represents an attributed graph, where $V$ is a set of vertices, $E\subseteq(V \times V)$ is a set of edges, $D_V$ and $D_E$ represent the dimension of node features and the number of edge types, respectively,  and $F_V$ is a $|V|\times D_V$ matrix of the features of all nodes, $F_E$ is a $|E|\times D_E$ matrix of the features of all edges, $F_G$ is the $D_G$-dimension global attributes of graph.}

\section{3 ~  Deep Graph Bayesian Optimization}

\subsection{3.1 ~ The DGBO Framework}
To tackle the two challenges mentioned in Section 1, we propose the DGBO framework based on BO. The DGBO poses the graph structure optimization
 as a sequential decision problem: which graph should be evaluated next so as to maximize the black-box $f$ as quickly as possible, by taking into account the information gain with uncertainty obtained from previous evaluations. There are two key components needed to be specified in the DGBO, i.e., a surrogate function and an acquisition function.

Surrogate function is used to approximate the block-box objective $f$.
GPs is the most popular one used in BO community due to its flexibility.
However, GPs suffers from the above-mentioned limitations especially its cubic complexity. Hence, in the DGBO we propose to use a deep graph neural network as the surrogate rather than GPs. Theoretically, it has been proved that the neural network with the infinite hidden layer is equivalent to GPs \cite{GPML2006}. Specifically, in order to automatically extract features from attributed graphs, we propose a novel surrogate architecture (see Fig.\ref{fig:DGBO_ipf_architecture3}) inspired by GCN (graph convolution networks) \cite{Bronstein2016Geometric}. Being a cutting-edge technique of network representation learning, GCN has succeeded in many graph-specific tasks \cite{kipf2017semi,Defferrard2016Convolutional,Schlichtkrull2017Modeling}.
In addition, to make our surrogate more scalable and be able to model uncertainty, we integrate a layer of BLR (Bayesian linear regressor) into the proposed deep graph neural network. It is worth noting that it is the introduction of BLR that makes the DGBO achieve linear complexity w.r.t the number of observations. The proposed surrogate model will be elaborated in the next section.

Acquisition function quantifies the potential of candidate graphs based on previous validations.
Given a graph search space $\mathcal{G}$ and a hyper-parameter space ${\Theta}$, acquisition function is defined as $\mathcal{U} : \mathcal{G}\times \Theta \to \mathbb{R}$.
The EI (expected improvement) \cite{Mockus1978The} is a commonly used criterion. Let ${\bm{\theta}}$ be hyper-parameters of the surrogate, the EI expresses
$
	\mathcal{U}(G|\mathcal{D}_t,{\bm{\theta}})=(\mu(G)-y_{max})\Phi(z(G))+\sigma(G)\phi(z(G)),
	\label{tab:EI}
$
where $z(G)=\frac{\mu(G)-y_{max}}{\sigma(G)}$,
$\mu(G)$ and $\sigma(G)$ are predictive mean and standard deviation (see Eq. \ref{eq:predicton_mu} and Eq. \ref{eq:predicton_sigma} for details),
 $y_{max}$ is the maximum value among current validations $\mathcal{D}_t=\{(G_1,y_1),(G_2,y_2),...,(G_t,y_t)\}$, $\Phi(.)$ and $\phi(.)$ denote the cumulative distribution function and probability density function of normal distribution, respectively.
Herein, we use the MCMC version of EI as described in \cite{DBLP:conf/nips/SnoekLA12}. For a fully-Bayesian treatment, this version integrates out hyper-parameters in the posterior distribution of observations, instead of point estimation which often causes local optimum. Thus, the final acquisition function is formulated as:
\begin{equation}
	\alpha(G|\mathcal{D}_t)=\int{\mathcal{U}(G|\mathcal{D}_t,{\bm{\theta}})}d{\bm{\theta}}\\
	\propto\sum_{i=1}^{S}{\mathcal{U}(G|\mathcal{D}_t,{\bm{\theta}}^{(i)})},
	\label{tab:equation_EIMCMC}
\end{equation}
where ${\bm{\theta}}^{(i)}\sim p({\bm{\theta}}|\mathcal{D}_t)$ and $p({\bm{\theta}}|\mathcal{D}_t)$ is the posterior distribution of ${\bm{\theta}}$, which will be discussed in Section 3.3.

By maximizing the above acquisition function, we can select a potential graph to evaluate next.
Then, one can recalculate the predictive mean and variance of surrogate based on previous validations and reselect next graph to be evaluated, until reaching a predefined termination condition. The framework of DGBO is given in Algorithm \ref{alg:Framwork}.
 \begin{algorithm}[!htb]
   \caption{DGBO}
   \label{alg:Framwork}
   \KwIn{Graph search space $\mathcal{G}$;
    The architecture of deep surrogate model $Net$;
    \# initialization evaluations $M$;
    Maximum \# iterations $MaxIter$;
    \# hyper-parameter sampling $S$;
    \# iterations of retraining $Re$;}
   \KwOut{The optimal graph $G^*$.}
   Initialize $M$ graphs randomly, evaluate them, and integrate into $\mathcal{D}_0$$=\{(G_1,y_1),(G_2,y_2)$$,...,(G_M,y_M)\}$;\\
   Train $Net$ with training set $\mathcal{D}_0$;\\
   Sampling $S$ hyper-parameter samples from their posterior distribution $p({\bm{\theta}}|\mathcal{D}_0)$;\\
   \For{$t = 1,2,...,MaxIter$}
   {
        Select a potential graph $G_{next}$  from $\mathcal{G}$ by maximizing Eq.\ref{tab:equation_EIMCMC} using random sampling;\\
        Evaluate the black-box system to obtain $y_{next}$, and augment data $\mathcal{D}_t=\mathcal{D}_{t-1}\cup \{(G_{next},y_{next})\}$;\\
        \If{$t\%Re==0$}
        {
            Retrain $Net$ with $\mathcal{D}_t$;
        }
        Resampling $S$ hyper-parameter samples from the posterior distribution $p({\bm{\theta}}|\mathcal{D}_t)$;\\
   }
   \Return $G^*$ with the maximum $y$ in $\mathcal{D}_{t}$.
 \end{algorithm}
%

\subsection{3.2 ~ The Proposed Deep Surrogate Model}
\begin{figure}[ht]
\centering
\includegraphics[scale=0.37]{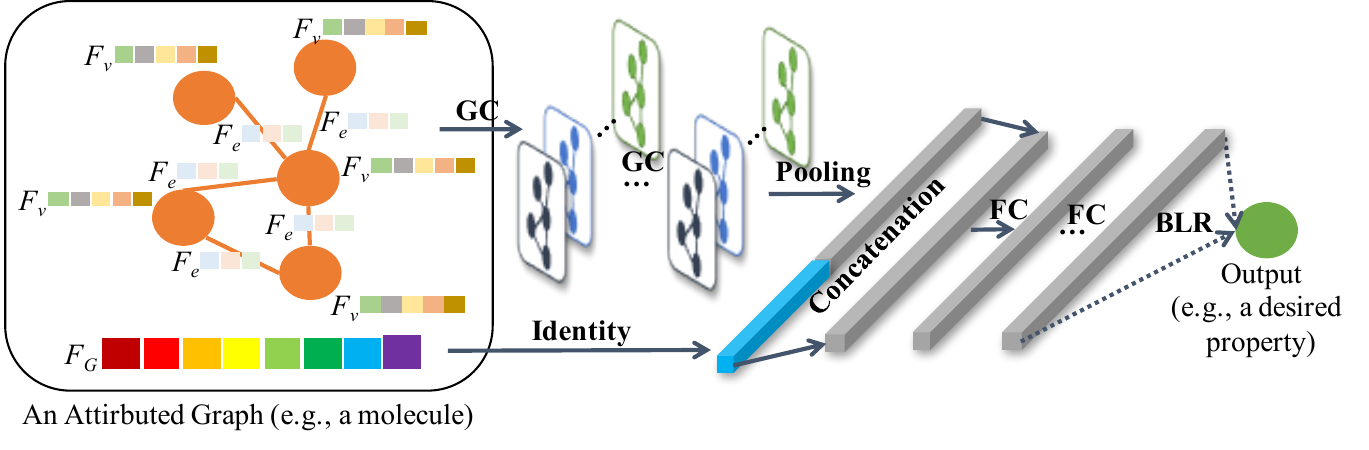}
\caption{The overview architecture of the surrogate model in the DGBO. Its input is an attributed graph (e.g., a molecular graph) and output is a continuous measure (e.g., a desired property). $F_v$ denotes the features of node $v$, $F_e$ denotes the features of edge $e$ and $F_G$ denotes global attributes.}
\label{fig:DGBO_ipf_architecture3}
\end{figure}

Fig.\ref{fig:DGBO_ipf_architecture3} shows the architecture of the proposed surrogate model. Each layer of the surrogate is discussed as follows.

\textbf{Graph convolution (GC) layer.}
To handle attributed graphs, we use graph convolution technique to automatically extract features of graphs without human intervention.

The existing works related to GC fall into three categories: spatial, spectral, and spectrum-free.
Since the surrogate of DGBO is used to approximate a graph regression function, we focus on designing a spectrum-free method, which actually is a polynomial approximation of spectral method.
Specifically, we propose a new spectrum-free convolution operation on attributed graphs, which is formulated as:
\begin{equation}
	H^{(l+1)}=\sigma(\sum_{r=1}^{D_E}\tilde{D}_{r}^{-1/2}\tilde{A}_{r}\tilde{D}_{r}^{-1/2}H^{(l)}W_{r}^{(l+1)}),
	\label{eq:graph_convolution_rel}
\end{equation}
where $H^{(l)}$ denotes the hidden representation of all nodes on layer $l$, $W_{r}^{(l+1)}$ denotes the weights on the type $r$ (of edges) at layer $l+1$, and $\sigma(.)$ denotes an activation function, such as $ReLU(.) = max(0, .)$ or $tanH$ function.
$\tilde{D}_{r}^{-1/2}\tilde{A}_{r}\tilde{D}_{r}^{-1/2}$ denotes the normalization representation of adjacent matrix $A_{r}$, where $\tilde{A}_{r}=A_{r}+I$, $\tilde{D}_{r}=D_{r}+I$, $D_{r}$ is a diagonal matrix and its diagonal elements are the degree of the corresponding nodes on type $r$. This normalization trick has been proved to be a first-order approximation of localized spectral filters on graphs.
Note that, unlike \cite{Schlichtkrull2017Modeling}, we utilize this trick instead of $D^{-1}_{r}A_{r}$ to normalize.

Compared with the current GC methods, our spectrum-free convolution model has the following advantages:
1) It is not necessary to manually design a fixed-size area of convolution operation on graph space.
2) Unlike spectral methods \cite{Bruna2014Spectral,Bronstein2016Geometric}, the parameters learned by different domains can be shared.
3) There are relatively fewer parameters to be trained. This is particularly desired for the BO-style optimization, where training data are usually sparse due to expensive validation cost. 4) Moreover, it is computationally time-saving to learn such parameters without the need to calculate eigen-system.

\textbf{Pooling layer.}
Through pooling layer, we wish to reasonably learn the global representation of the whole graph from the local representations of nodes and edges.
Accordingly, we propose the following pooling operation:
\begin{equation}
	H^{(pool)}=\sigma(sum_{row}(softmax(H^{(l)}W^{(pool)}))),
	\label{eq:pooling}
\end{equation}
where $H^{(pool)}$ denotes the representation of graph-level features, $H^{(l)}$ denotes the output after going through $l$ graph convolution layers, and $W^{(pool)}$ denotes the weights on pooling layer. We first multiply $H^{(l)}$ by $W^{(pool)}$ to map the features obtained by graph convolution into a new latent space with a specified dimension, and then  apply $softmax(.)$ to the result to obtain row-wise sparse representations and map them into a unified interval $[0,1]$. Note that the usage of $softmax(.)$ can also prevent ignoring important rows but having some relatively small-value dimensions. Then we apply $sum_{row}(.)$ to accumulate multiple row-wise features into one vector before applying a nonlinear function $\sigma(.)$ to obtain final graph-level representation.

\textbf{Prior layer.}
The prior knowledge about a graph (such as the weight of a molecule) can be regarded as the global attributes of the graph. In addition to prior knowledge, the global information (such as the scale of a graph) or the high-order structural information (such as betweenness centrality or clustering coefficient) that are easy to get can also be regarded as the global attributes of a graph. All of the available information could potentially promote the performance of optimization. Through the prior layer, such global attributes are expected to be logically integrated into final graph representation. There might be different ways to do this. In this work, we adopt a simple concatenating strategy in order to make optimization as scalable as possible,
\begin{equation}
	H^{(con)}=Concat(H^{(pool)},\lambda F_G),
	\label{eq:concatenation_operation}
\end{equation}
where $F_G$ denotes the global attributes of graph, $H^{(pool)}$ denotes the representation output by pooling layer, and $\lambda$ is a switch weight. If global attributes are available, we turn $\lambda$ on. Otherwise, we turn it off.

\textbf{Bayesian linear regressor (BLR) layer.}
To predict the measure of a graph while capturing uncertainty, we add a Bayesian linear regressor (BLR) just behind multiple FC (fully connected) layers as the last layer of the surrogate architecture.
We regard this model as an adaptive basis regression and the basis functions are parameterized by the weights and biases of the deep neural network. BLR is formulated as:
\begin{equation}
	\bm{y}_{1:N}=\Phi(.)^T\bm{w}+\bm{b},
	\label{eq:BLR}
\end{equation}
where $\bm{y}_{1:N}$ denotes outputs, $\bm{b}\sim\mathcal{N}(\bm{0},\sigma_{noise}^2\bm{I})$, $\mathcal{N}(.)$ is normal distribution, $\sigma_{noise}^2$ is noise level, $\bm{w}$ is the weights of BLR layer, and $\Phi(.)$ is the decision matrix output by previous layers as the input of BLR layer.
Given a prior distribution on weights: $\bm{w}\sim \mathcal{N}(\bm{0},\sigma_{\bm{w}}^2\bm{I})$, where $\sigma_{\bm{w}}^2$ denotes the uncertainty of $\bm{w}$, the measure of $G^*$ can be predicted by:
\begin{equation}
	y^*|\mathcal{D}_{1:N},\bm{y}_{1:N},G^*\sim \mathcal{N}(\mu(G^*),\sigma^2(G^*)),
	\label{eq:predicton}
\end{equation}
where $\mathcal{D}_{1:N}$ are observations, $\bm{y}_{1:N}$ are evaluated  measures,
\begin{equation}
	\mu(G^*)=\sigma_{noise}^{-2}\Phi(G^*)^TK^{-1}\Phi(.)\bm{y}_{1:N},
	\label{eq:predicton_mu}
\end{equation}
\begin{equation}
	\sigma^2(G^*)=\Phi(G^*)^TK^{-1}\Phi(G^*)+\sigma_{noise}^2.
	\label{eq:predicton_sigma}
\end{equation}

\noindent $K=\sigma_{noise}^{-2}\Phi(.)\Phi(.)^T+\sigma_{\bm{w}}^{-2}\bm{I}$.
Note that this layer is the key to reduce time complexity (see Section 3.4 for details).

\textbf{Loss function.}
 Having each layer of the deep surrogate model, we use the following loss function to train it,
$
	loss=\sum_{i=1}^{N}|\hat{y}_i-y_i|^2+\gamma ||\bm{\Omega}||_{l_2}^2,
	\label{eq:loss_function}
$
where $\hat{y}_i$ denotes predictive output, $y_i$ denotes ground truth, $\bm{\Omega}$ denotes the weights and biases of neural network, and $\gamma$ denotes penalty coefficient.

\subsection{3.3 ~  Implementation Details}
\label{sec_implementation_details}
\textbf{Handling hyper-parameters.}
In the proposed deep surrogate model, all hyper-parameters need to handle include $\bm{\Omega}$ in loss function (the weights and biases of GC, pooling and FC layers) and $\bm{\theta}$ in Eq. \ref{tab:equation_EIMCMC} ($\sigma_{\bm{w}}^2$ and $\sigma_{noise}^2$ of BLR layer).
For $\bm{\Omega}$, we train GC, pooling and FC layers via backpropagation and a wildly used stochastic gradient descent named Adam \cite{Kingma2015Adam}.
In this training phase, we use a linear output layer to replace BLR. This process can be viewed as a MAP estimate of all parameters in these layers.
Based on the parameterized basis functions, we make predictions by a BLR.
Thereby we need to deal with $\bm{\theta}$ of the BLR layer.
For a full-Bayesian treatment, we integrate out $\bm{\theta}=\{\sigma_{\bm{w}}^2, \sigma_{noise}^2\}$ by using an ensemble MCMC sampler \cite{2013PASP}, according to their posterior distribution $p({\bm{\theta}}|\mathcal{D})$.
For the posterior distribution,
we place a logarithmic normal prior with mean -10 and standard deviation 0.1 on $\sigma_{\bm{w}}^{-2}$, and a horseshoe prior with scale 0.1 on $\sigma_{noise}^2$.
Then, the posterior is obtained by $p({\bm{\theta}}|\mathcal{D})\propto p(\mathcal{D}|{\bm{\theta}})p(\bm{\theta})$, where $p(\mathcal{D}|{\bm{\theta}})$ is the marginal likelihood of evaluations.

\textbf{Basis regularization.}
Note that, in Eq. \ref{eq:graph_convolution_rel}, the number of parameters $W_{1},W_{2},...,W_{D_E}$ will increase rapidly with the number of edge types $D_E$ on each GC layer. Hence we adopt \textit{basis regularization} to prevent overfitting by reducing the number of parameters.
Basis regularization assumes that different relations may partially share common parameters.
Specifically, we assume that $W_r^{(l+1)}$ consists of a linear combination of bases $\{V_1^{(l+1)},V_2^{(l+1)},...,V_B^{(l+1)}\}$, i.e.,
$W_r^{(l+1)}=\sum_{b=1}^{B}\beta_{r,b}^{(l+1)}V_{b}^{(l+1)}$, where $\beta$ is combination coefficient and $B$ is the number of bases.

\textbf{Optimizing surrogate architecture by transfer.}
The architecture of deep surrogate model in Fig.\ref{fig:DGBO_ipf_architecture3} will affect the efficacy of the DGBO.
To design a proper surrogate architecture is a key step to further accelerate the optimization process. One promising way is to optimize this architecture based on the data of the task in question.
However, in the real world, we often have very limited observations for the task at hand due to the high cost of function evaluation.
To address this issue, in the paper we suggest employing the idea of transfer learning, i.e., to optimize surrogate architecture based on the available data from other sources.
Specifically, we represent surrogate architecture as a 11-dimension vector (see Table \ref{tab:optimal_surrogate_architecture}).
The performance of architecture is defined as the regression accuracy on a molecular data set CEP, which includes $\sim$20k organic molecules and their photovoltaic
efficiency, which contributed by Harvard Clean Energy Project \cite{Hachmann2011The}. We randomly selected 1,000 molecules from CEP, among which 500 for training and 500 for testing.
Then, we use a GPs-based BO to optimize architecture based on CEP. The optimal architecture obtained in this way is
shown in Table \ref{tab:optimal_surrogate_architecture}. We will apply the architecture to all tasks discussed in our experiments.
\begin{table}[!htb]
	\small
	\centering
	\begin{tabular}{ccc}\hline
		Parameters&Ranges&Optimal\\\hline
		\# GC layers&\{1, 2, 3, 4, 5\}&5\\
		\# FC layers&\{1, 2, 3, 4, 5\}&5\\
		\# units of GC&[10, 100]&48\\
		\# units of pooling&[10, 100]&50\\
		\# units of FC&[10, 100]&45\\
		$\sigma(.)$ of GC&\{$ReLU$, $tanH$\}&$tanH$\\
		$\sigma(.)$ of pooling&\{$Identity$, $ReLU$, $tanH$\}&$Identity$\\
		$\sigma(.)$ of FC&\{$Identity$, $ReLU$, $tanH$\}&$tanH$\\
		Learning rate&[1e-4, 1e-1]&1e-4\\
		Dropout&[0, 1]&0.0\\
		Penalty coefficient&[1e-5, 1e-1]&1e-5\\\hline
	\end{tabular}
	\caption{The optimal surrogate architecture.}
	\label{tab:optimal_surrogate_architecture}
\end{table}

\subsection{3.4 ~  Time Complexity Analysis}

\begin{figure}[!htb]
\centering
\includegraphics[scale=0.35]{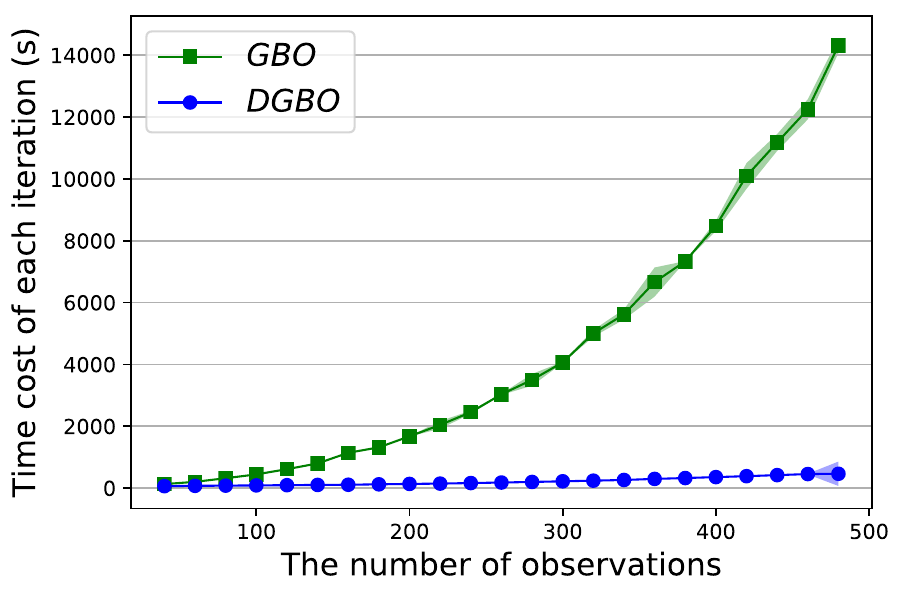}
\caption{Comparison of scalability between the DGBO and a non-scalable method GBO on Synthetics (described in section 4.1). $x$-axis and $y$-axis denote \# validations and the time cost of each iteration, respectively.
The time cost of each iteration includes the time of selecting next graph plus the time of retraining deep surrogate network in the DGBO or learning hyper-parameters of kernels in GBO.
}
\label{fig:time_cost}
\end{figure}

In BO-style optimization, maximizing acquisition function is often the bottleneck of efficiency. If the surrogate is GPs, it will take $O(N^3)$ time to compute the inverse of an $N$-by-$N$ kernel matrix and then, based on it, to predict the mean and variance required by acquisition function. $N$ is the number of validations. Similarly, in the DGBO, maximizing acquisition function (step 5 in Algorithm 1) is still the most expensive relative to others (including training deep surrogate in step 8), which dominates the overall time of the DGBO. We will see, by using the proposed deep surrogate, the time for quantifying acquisition function will be greatly reduced to linear order.
Let $M$ be the number of units on BLR layer. The matrices $\Phi(\cdot)$ and $K$ in Eqs. \ref{eq:predicton_mu} and  \ref{eq:predicton_sigma} are $N$-by-$M$ and $M$-by-$M$, respectively. It takes $O(M^2N)$ and $O(M^3)$ time to compute $K$ and its inverse. It then takes $O(M^2N)$ and $O(M^2)$ time to compute Eq. \ref{eq:predicton_mu} (mean) and  Eq. \ref{eq:predicton_sigma} (variance). Note $M$ is a constant much less than $N$. So the total time of prediction is $O(M^2N)=O(N)$. Moreover, we empirically validate this by comparing the DGBO to a GPs-based method (GBO) \cite{Cui2018}. It can be seen from Fig.\ref{fig:time_cost} the DGBO increases linearly with the number of validations, far superior to the cubic growth of the GBO.

\section{4 ~ Experiments}
\label{sec_experiments}
Here, we rigorously evaluate the DGBO by answering three questions.
1) Can the available features from attributed graphs benefit optimization?
2) How effective and efficient is the DGBO compared with the start-of-the-art on real-world problems?
3) Can it be applied to various domains?

\subsection{4.1 ~ Data Sets}
The data sets used in this paper are summarized in Table \ref{tab:data_sets}.
\begin{table}[ht]
	\small
	\centering
	\begin{tabular}{cccccccc}\hline
		Data sets&$|\mathcal{G}|$&$\overline{|V|}$&$\overline{|E|}$&$D_V$&$D_E$&$D_G$\\\hline
		Synthetics&500&39.8&141.5&-&-&2/4/6\\\hline\hline
		Delaney&1,122&13.3&27.4&68&6&6\\
		ZINC&20,000&23.2&24.9&68&6&6\\\hline\hline
		SiouxFalls&32,768&24&30.5&2&-&5\\\hline
	\end{tabular}
	\caption{Statistics of data sets. $|\mathcal{G}|$ denotes the graph numbers of search space, $\overline{|V|}$ denotes the average number of nodes, and $\overline{|E|}$ denotes  the average number of edges.}
	\label{tab:data_sets}
\end{table}

\textbf{Synthetics} is artificially generated via the NetworkX tool \cite{SciPyProceedings_11}, which includes 500 undirected random graphs.
Note that both the nodes and edges of graphs in this data set do not have features.
For the global attributes, we extract 6 features from each graph: \# nodes
$x_1$, \# edges $x_2$, average degree centrality $x_3$, average betweenness centrality
$x_4$, average clustering coefficient $x_5$, and a completely unrelated random variable $x_6$.

\textbf{Delaney} is a molecular data set having 1,122 molecules whose aqueous solubility has been measured by \cite{Delaney2004ESOL}.
Node feature is the atomic type, edge feature is the chemical bond type, and each molecule has six additional global attributes (see Fig.\ref{fig:mol2graph}).

\textbf{ZINC} includes 20,000 drug-like commercially available molecules extracted at random from the ZINC database \cite{Irwin2012ZINC}.
The features of node and edge are the same as Delaney.
Since ZINC does not provide global attributes for each molecule, we extract some structure information including \# nodes, \# edges, average degree centrality, average betweenness centrality, average closeness centrality, and average clustering coefficient as global attributes.

\textbf{SiouxFalls}. This data set is widely used in transportation studies \cite{Leblanc1975An}.
Similar to the previous works, we randomly remove a number of roads from the original network, by assuming these roads have not been built yet. We now want to decide which roads of these removed roads should be constructed in order to minimize total travel time. For example, if we remove 15 roads,  there will be $2^{15}$ potential assignments, i.e., our search space will contain total 32,768 candidate road graphs. The origin-destination matrix used in our simulations is the same as \cite{Leblanc1975An}.
Node feature is a two-dimensional continuous coordinate of the intersection and global attributes are the same as the ones used in ZINC.

\begin{table*}[htb]
	\small
	\centering
	\begin{tabular}{ccccc}\hline
		Methods&Situation(a)(\# Evals=90)&Situation(b)(\# Evals=100)&Situation(c)(\# Evals=100)&Situation(d)(\# Evals=120)\\\hline
		Random&$2.796\pm0.004$&$2.796\pm0.004$&$2.796\pm0.004$&$2.796\pm0.004$\\
		BO$_{vec}$&\bm{$2.863\pm0.000$}&$2.824\pm0.002$&$2.828\pm0.005$&$2.851\pm0.000$\\
		BO$_{Glets}$&$2.814\pm0.005$&$2.834\pm0.001$&$2.834\pm0.001$&$2.850\pm0.000$\\
		BO$_{dGlets}$&$2.815\pm0.002$&$2.815\pm0.002$&$2.815\pm0.002$&$2.821\pm0.001$\\
		GBO$_{dGlets}$&\bm{$2.863\pm0.000$}&\bm{$2.863\pm0.000$}&\bm{$2.863\pm0.000$}&$2.849\pm0.000$\\\hline\hline
		DGBO$_{noRel}$&\bm{$2.863\pm0.000$}&\bm{$2.863\pm0.000$}&\bm{$2.863\pm0.000$}&\bm{$2.863\pm0.000$}\\\hline
	\end{tabular}
	\caption{Evaluation of the DGBO versus Random and other non-scalable methods on Synthetics in four situations.
	\# Evals represents the evaluation times.
	The mean and standard deviation of $y$ are reported.
	Bold positions are the optimums.
	}
	\label{tab:Comparison_accuracy3}
\end{table*}

\subsection{4.2 ~ Baselines and Setup}
The baseline methods compared to the DGBO are categorized into three groups as follows.
\begin{itemize}
\item Random denotes random selection at each iteration.
\item BO denotes GPs-based BO method integrating different kernels indicated by the subscript. The specific kernels include Gaussian ARD kernel (BO$_{vec}$), graphlets kernel \cite{Shervashidze2009Efficient} (BO$_{Glets}$), and deep graphlets kernel \cite{Yanardag2015Deep} (BO$_{dGlets}$).
As the Gaussian ARD kernel is a vectorial kernel, we firstly extract the features of graph manually, and then apply it to these pre-extracted hand-crafted features.
\item GBO denotes the GPs-based BO method by combining a graph kernel and a vectorial kernel \cite{Cui2018}. Subscript indicates its integrated graph kernel.
The specific kernels include deep graph kernels based on graphlets (GBO$_{dGlets}$), subtree patterns (GBO$_{dWL}$) and shortest path (GBO$_{dSP}$), respectively.
Note that the first kernel can deal with \textit{unlabelled} graphs while the last two kernels can only deal with \textit{labelled} graphs.
\end{itemize}
For the proposed method, DGBO$_{noRel}$ denotes that it ignores edge features (i.e., applies Eq. \ref{eq:graph_convolution_rel} in which $D_E\equiv1$ to convolute graph),
DGBO$_{Rel}$ denotes that it considers edge features via Eq. \ref{eq:graph_convolution_rel} without basis regularization,
and DGBO$_{RelReg}$ denotes that it not only considers edge features, but also uses basis regularization.
Without specification, both the number of initializing graphs $M$ and the iterations of retraining $Re$ are set to 20, $B$ is set to 4, and
all algorithms run 5 times to eliminate random effects.

\subsection{4.3 ~ Artificial Non-Linear Function}
We firstly test the efficacy of the DGBO on Synthetics.
Note that there are no features on nodes in this data set. In this case, a common way to assign features to nodes is: the nodes of the same graph are assigned the same one-hot representation, in which the entry of {1} corresponds to the index of the graph.
We, then, normalize each global attribute into [0,1] via $\tilde{x}=\frac{x-x_{min}}{x_{max}-x_{min}}$.
We define the target $y=-Hart(\tilde{x}_1,\tilde{x}_2,\tilde{x}_3,\tilde{x}_4)$ as the artificial non-linear function from a graph to a functional measure,
where $Hart(.)$ denotes the four-dimension Hartmann function that is a common non-linear test function in BO community.
We test the DGBO to find a graph with the maximum $y$ from Synthetics in four situations:
(a) properly using $\tilde{x}_1$, $\tilde{x}_2$, $\tilde{x}_3$, and $\tilde{x}_4$ as global attributes;
(b) partially using $\tilde{x}_1$ and $\tilde{x}_2$ as global attributes;
(c) totally using $\tilde{x}_1$, $\tilde{x}_2$, $\tilde{x}_3$, $\tilde{x}_4$, $\tilde{x}_5$ and $\tilde{x}_6$ as global attributes;
(d) falsely using a non-direct related feature $\tilde{x}_5$ and a completely unrelated feature $\tilde{x}_6$ as global attributes.

In Table \ref{tab:Comparison_accuracy3},
no matter how the prior knowledge is changed from the situation (a) to (d), the DGBO outperforms other non-scalable methods.
Note that the DGBO outperforms BO$_{Glets}$ and BO$_{dGlets}$ significantly.
That implies the graph convolution used in the DGBO is more suitable for attributed graph feature extraction than the existing graph kernels.
Moreover, when the global features pre-extracted by hand are unrelated to the objective in question (i.e., situation
(d)),  all compared methods fail to find the optimum mainly because these handcrafted features may impose a negative effect on optimization
process. To prevent overfitting is another reason why a simple concatenation fusion rather than a more complicated one is preferred on the prior layer.
 Note GBO$_{dWL}$ and GBO$_{dSP}$ are not included as both cannot handle the graphs in which all nodes have identical labels.

\subsection{4.4 ~ Molecular Discovery}
Molecular discovery is a meaningful problem.
However, the optimization
in molecular space is extremely challenging because the search space is usually large, and molecules have a rich set of available features (see an illustration in Fig.\ref{fig:mol2graph}). More importantly, it is very expensive to evaluate a molecule by doing regardless of simulations or real experiments.
Thus, the DGBO is suitable for this problem.
Specifically, we apply it to discover optimal molecules from two graph spaces of Delaney and ZINC, respectively.

\begin{figure}[!htb]
\centering
\includegraphics[scale=0.36]{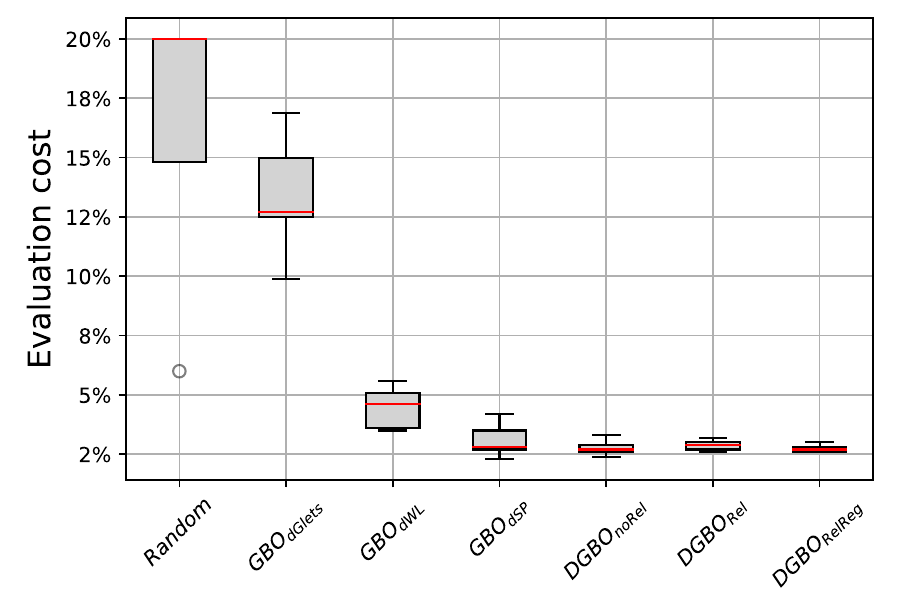}
\caption{Boxplot of evaluation cost for finding the optimum by the DGBO versus other baselines on Delaney.
$y$-axis indicates the percentage of evaluated graphs over all candidate graphs in search space.
}
\label{fig:exp2-Discovery_of_Molecules-delaney-box3}
\end{figure}

In Delaney, we aim to find a molecule with maximal aqueous solubility.
All methods can find the optimum under a given evaluation budget (i.e., 200), except for Random.
In Fig.\ref{fig:exp2-Discovery_of_Molecules-delaney-box3},
we see that all model-based methods outperform Random.
GBO$_{dGlets}$ is significantly worse than other model-based methods, as it cannot use node features (i.e., atomic type).
Note that our methods outperform others, i.e., they only evaluate about 3\% of the whole search space to find the optimum.
Meanwhile, they are more robust to initial validations.
Moreover, DGBO$_{Rel}$ and DGBO$_{RelReg}$ are slightly more stable than DGBO$_{noRel}$, as they take advantage of edge features (i.e., chemical bond type).

To further test the efficacy and scalability of the DGBO on a larger search space, we apply it to ZINC to find an optimal drug-like molecule with maximal $y=5\times QED-SAS$, where $QED$ denotes the quantitative estimation of drug-likeness \cite{Bickerton2012Quantifying} and $SAS$ denotes the synthetic accessibility score \cite{Ertl2009Estimation}. That is, we want to find the most drug-like molecule that is also easy to synthesize.
In addition, we compare the DGBO with a state-of-the-art technology of automatic chemical design, named as VAE+GPs \cite{G2018Automatic}.
The process of the VAE+GPs is as follows:
a variational autoencoder (VAE) is firstly trained upon whole ZINC database to map all molecules into a fixed-length (e.g., 196) continuous vector space, and then a GPs-based BO is used to find optimal molecules in this latent space.

\begin{figure}[!htb]
\centering
\includegraphics[scale=0.39]{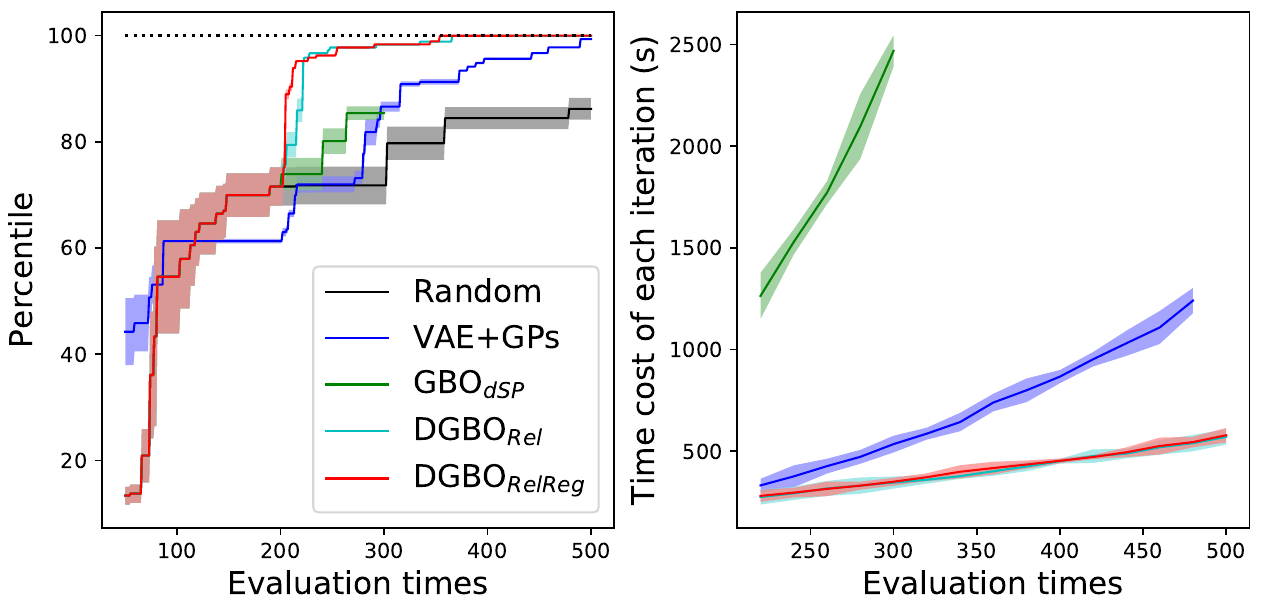}
\caption{Left: Comparison of convergence curves on ZINC. $x$-axis represents evaluation times. $y$-axis represents the percentile of optimum.
We randomly evaluated 200 molecules at the initialization stage and ran all methods on the same hardware setting.
Solid lines represent mean values, and shaded regions represent variance.
Right: Time cost of each iteration. $y$-axis represents the time cost of each iteration.}
\label{fig:zinc}
\end{figure}

From Fig.\ref{fig:zinc} we see the DGBO outperforms others significantly and needs the
minimal time cost. The DGBO finds the optimum by evaluating only 1.8\% of whole search space. DGBO$_{RelReg}$ is slightly better than DGBO$_{Rel}$, which shows the efficacy of basis regularization.
On the other hand, VAE+GPs needs much more evaluations to find a near optimum, and its scalability with the number of evaluations is remarkably worse than that of the DGBO. In addition to using expensive GPs, the representation of the graph adopted by it might be another reason for inefficiency. VAE+GPs learns graph representation
via an unsupervised manner, which may not be insightful for specific tasks. Note that GBO performs poorly
because it has to stop much earlier than convergence due to its prohibitively high time cost.

\subsection{4.5 ~ Urban Road Network Design}
\begin{figure}[!htb]
\centering
\includegraphics[scale=0.32]{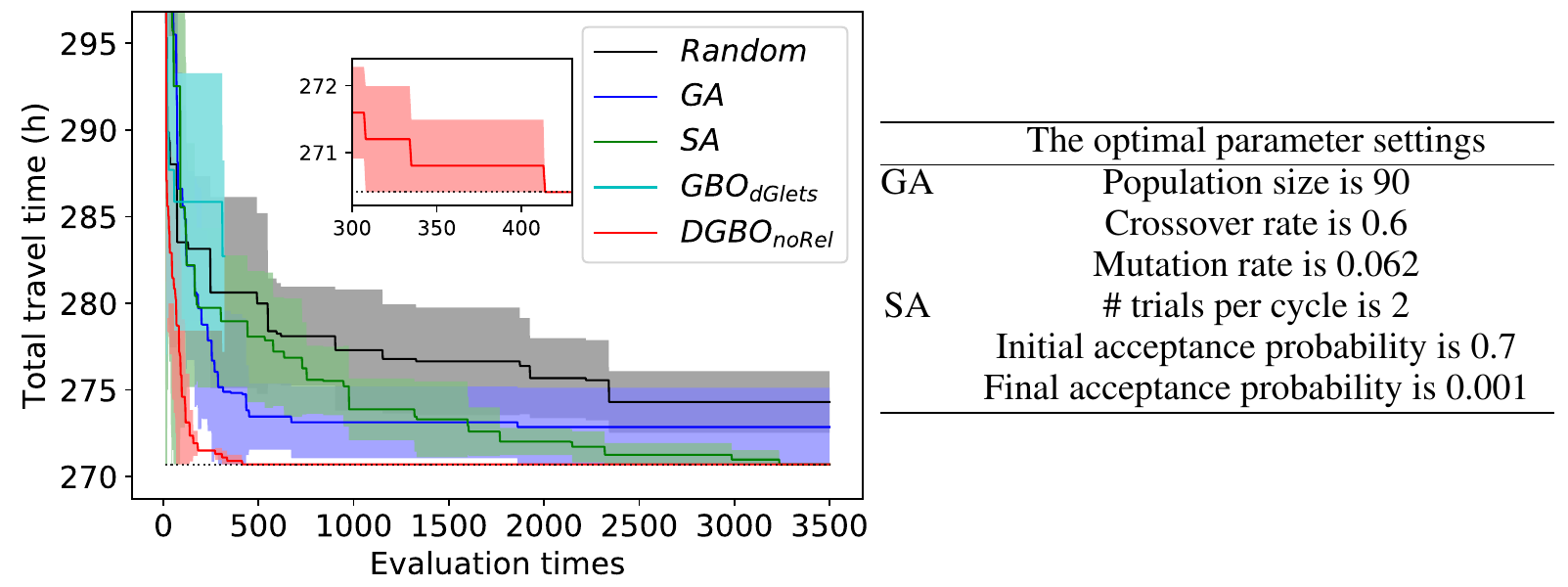}
\caption{Left: Convergence comparison of respective methods on urban road network design.
Right: The optimal parameters of GA and SA.
}
\label{fig:exp-traffic}
\end{figure}

In order to verify the effectiveness of the DGBO in different domains,
we apply it to address the task of urban road network design \cite{Farahani2013A} on SiouxFalls.
Urban road network design is a bi-level optimization problem.
The upper-level problem concerns global policy design
in practice which aims to achieve an optimal macroscopic measure (e.g., reducing total traveling time) by designing new policies  (e.g.,
where to build new roads). While, the low-level problem cares about how to optimize the behaviors of individuals, e.g., the distribution of traffic flow in a given road net.
Herein, we focus on the upper-level road network design problem, and for the lower-level problem,
we use the Frank-Wolfe algorithm \cite{FUKUSHIMA1984169}, a widely used method in transportation, to optimally distribute traffic flows.
However, this way usually takes expensive computing resources, particularly for very large road nets. Thus, how to design an optimal road network under a few evaluations is still a challenging problem.
In addition, according to \cite{Farahani2013A}, genetic algorithm (GA) \cite{Xiong1992Transportation} and simulated annealing (SA) \cite{Miandoabchi2011Optimizing} are two most common optimization algorithms for this problem.
Therefore, we compare the DGBO against GA and SA as baselines.
Moreover, we use a GPs-based BO to optimize the parameters of GA and SA in order to achieve their best performance (see the right panel of Fig.\ref{fig:exp-traffic}).

From Fig.\ref{fig:exp-traffic} we see that the DGBO significantly outperforms others again.
It finds the optimum under less than 420 evaluations, which is nearly 7.7 times faster than the SA ($\sim$3250 evaluations), which runs the second fastest.
The main reason is that the DGBO can take advantage of both structure and node features, while the SA and GA cannot.
Note GBO stops very early again due to its high time cost.

\section{5 ~ Conclusions}
\label{sec_conclusions}

In this work, we propose the DGBO, a novel scalable global optimization method on attributed graphs.
To rigorously test its effectiveness, we apply the DGBO to solve various problems.
The results show that the DGBO significantly outperforms the state-of-the-art methods in terms of both accuracy and scalability.
Based on this work, the scalability of the proposed framework can be further enhanced through parallelization for those problems involving much larger search spaces, such as neural architecture search.

\section{ Acknowledgments}
This work was supported by the National Natural Science Foundation
of China under grants 61572226 and 61876069, and Jilin Province Key Scientific and
Technological Research and Development Project under grants 20180201067GX and 20180201044GX.

\fontsize{9.0pt}{10.0pt}
\selectfont
\bibliography{AAAI-CuiJ}
\bibliographystyle{aaai}

\end{document}